\documentclass{article} 
\usepackage[utf8]{inputenc}
\usepackage[preprint]{colm2026_conference}

\usepackage{microtype}
\usepackage{hyperref}
\usepackage{url}
\usepackage{booktabs}
\usepackage{amsmath, amssymb}
\usepackage{graphicx}
\usepackage{lineno}

\definecolor{darkblue}{rgb}{0, 0, 0.5}
\hypersetup{
    colorlinks=true,
    citecolor=darkblue,
    linkcolor=darkblue,
    urlcolor=darkblue
}

\title{Care-Conditioned Neuromodulation for\\
Autonomy-Preserving Supportive Dialogue Agents}

\author{
Shalima Binta Manir\textsuperscript{1}, Tim Oates\textsuperscript{2} \\
Department of Computer Science \\
University of Maryland, Baltimore County \\
\texttt{smanir1@umbc.edu, oates@cs.umbc.edu}
}

\begin{document}

\ifcolmsubmission
\linenumbers
\fi

\maketitle

\begin{abstract}
Large language models deployed in supportive or advisory roles must balance helpfulness with preservation of user autonomy, yet standard alignment methods primarily optimize for helpfulness and harmlessness without explicitly modeling relational risks such as dependency reinforcement, overprotection, or coercive guidance. We introduce Care-Conditioned Neuromodulation (CCN), a state-dependent control framework in which a learned scalar signal derived from structured user state and dialogue context conditions response generation and candidate selection. We formalize this setting as an autonomy-preserving alignment problem and define a utility function that rewards autonomy support and helpfulness while penalizing dependency and coercion. We also construct a benchmark of relational failure modes in multi-turn dialogue, including reassurance dependence, manipulative care, overprotection, and boundary inconsistency. On this benchmark, care-conditioned candidate generation combined with utility-based reranking improves autonomy-preserving utility by +0.25 over supervised fine-tuning and +0.07 over preference optimization baselines while maintaining comparable supportiveness. Pilot human evaluation and zero-shot transfer to real emotional-support conversations show directional agreement with automated metrics. These results suggest that state-dependent control combined with utility-based selection is a practical approach to multi-objective alignment in autonomy-sensitive dialogue.
\end{abstract}

\section{Introduction}

Large language model (LLM) assistants are increasingly deployed in emotionally 
salient contexts such as mental health support, education, caregiving, and 
advisory interactions. In these settings, failures can occur not only through 
harmful content but through relational harms such as reinforcing reassurance-seeking, 
encouraging dependence, over-intervening, or applying pressure under the guise of care. 
These behaviors are not captured by standard safety benchmarks, which focus primarily 
on toxicity and dangerous content rather than long-term relational dynamics 
\citep{bender2021dangers, gabriel2020artificial}.

Existing alignment methods optimize for helpfulness and harmlessness using supervised 
fine-tuning, RLHF, or preference optimization \citep{ouyang2022training, bai2022constitutional, 
rafailov2023direct}, but do not explicitly model relational risks such as dependency, coercion, or overprotection. We therefore study a more specific alignment objective, \emph{support without dependency}: dialogue agents should provide support while preserving user agency, respecting boundaries, and encouraging independent coping.

We propose \textbf{Care-Conditioned Neuromodulation (CCN)}, a state-dependent control framework in which a learned scalar care signal conditions response generation and 
candidate selection based on structured user state and dialogue context. We also define 
an autonomy-preserving utility that balances autonomy support and helpfulness against 
dependency and coercion risk, and construct a benchmark of multi-turn supportive 
dialogue scenarios covering relational failure modes not captured by existing datasets 
such as ESConv and EmpatheticDialogues \citep{liu2021towards, rashkin2019towards}.

Empirically, care-conditioned candidate generation combined with utility-based reranking 
improves autonomy-preserving utility over supervised fine-tuning and preference 
optimization while maintaining comparable supportiveness, with additional support from 
pilot human evaluation and zero-shot transfer to real emotional-support conversations.
\paragraph{Conceptual inspiration.}
The term \emph{Care-Conditioned Neuromodulation} is inspired by Churchland's account of 
caring as a biologically grounded, state-dependent expansion of concern beyond the self 
\citep{churchland2011braintrust, churchland2019conscience}. Analogously to neuromodulatory 
signals that adjust neural circuit gain without rewriting synaptic weights, $m_t$ modulates 
decoding behaviour by shifting temperature and sampling diversity without modifying the 
underlying language model. When vulnerability or relational risk is elevated, $m_t$ shifts 
the response policy through state-conditioned decoding and utility-based selection. 
We do not claim biological fidelity.

\paragraph{Contributions and main findings.}
The full CCN pipeline care-conditioned generation combined with utility-based 
reranking improves autonomy-preserving utility by $+0.25$ over supervised 
fine-tuning and $+0.07$ over preference optimization while maintaining comparable 
supportiveness. Ablation results indicate that care-conditioned generation and 
utility-based selection play complementary roles, with the care signal contributing 
useful candidates for reranking.

The contributions are as follows:
\begin{itemize}
    \item We formulate supportive dialogue as a multi-objective alignment problem involving autonomy support, dependency risk, coercion risk, and supportiveness.
    \item We introduce Care-Conditioned Neuromodulation (CCN), a state-dependent control framework that conditions generation on structured user state and relational context.
    \item We construct a benchmark of relational failure modes in multi-turn supportive dialogue with structured user state and relational safety labels.
    \item We show that utility-based reranking over care-conditioned candidates improves autonomy-preserving utility over supervised fine-tuning and preference optimization, with supporting evidence from automated evaluation, human ratings, and transfer to real emotional-support conversations.
\end{itemize}

\section{Related Work}

Our work relates to supportive dialogue, LLM alignment, controllable generation, 
and long-horizon dialogue systems, with a focus on relational safety and autonomy preservation.

\subsection{Supportive and Emotional Dialogue}

Prior work has studied dialogue systems for empathy and emotional support, including 
EmpatheticDialogues and ESConv \citep{rashkin2019towards, liu2021towards}, as well as 
support strategy modeling and emotionally informed response generation 
\citep{majumder2020mime, smith2020can}. These systems primarily optimize for empathy 
and conversational quality rather than long-term autonomy preservation. In contrast, 
we study supportive dialogue as an autonomy-preserving alignment problem and explicitly 
model relational risks such as dependency and coercion.

\subsection{Alignment, Controllability, and Adaptation in LLMs}

Modern alignment methods rely on supervised fine-tuning, reinforcement learning from 
human feedback, and preference optimization \citep{ouyang2022training, bai2022constitutional, 
rafailov2023direct}. Controllable generation and parameter-efficient fine-tuning methods 
such as LoRA enable behavior steering without full retraining \citep{brown2020language, 
dathathri2019plug, hu2022lora, dettmers2023qlora}. Our approach differs in that we use a 
learned control signal representing user vulnerability to coordinate decoding and response 
selection for autonomy-preserving supportive dialogue.

\subsection{Memory and Stateful Dialogue}

Long-horizon dialogue systems use memory and retrieval to maintain user information across 
turns \citep{weston2014memory, lewis2020retrieval, zhang2018personalizing}. Prior work 
primarily stores factual or persona information, whereas our approach tracks relational state 
such as boundaries, commitments, and vulnerability.

\subsection{Relational Harms and Dependency in AI Systems}

Concerns about manipulation, dependency, and social influence in AI systems have been 
discussed in prior work \citep{bender2021dangers, gabriel2020artificial}. We contribute by 
operationalizing relational harms as measurable objectives and framing supportive dialogue as 
a multi-objective alignment problem.

A more detailed discussion of related work is provided in Appendix~C.

\section{Problem Setting: Support Without Dependency}

We consider a dialogue agent interacting with a user over multiple turns $t = 1, 2, \dots$. At each turn, the system observes the dialogue context $x_t$ and a structured user state:
\[
S_{d,t} = \{\text{goals, boundaries, preferences, vulnerability, commitments, stress context}\}.
\]

The objective is to provide supportive and protective responses while preserving user autonomy and avoiding harmful relational dynamics such as reassurance dependence, over-intervention, or coercive guidance. We refer to this objective as \emph{support without dependency}: the agent should assist the user while maintaining agency, encouraging independent coping, and respecting stated boundaries.

\subsection{Autonomy-Preserving Utility}

We formalize this objective using a utility function over candidate responses $y$ at turn $t$:
\[
U(x_t, y) = \lambda_1 V_{\text{aut}}(x_t, y)
- \lambda_2 Q_{\text{dep}}(x_t, y)
- \lambda_3 Q_{\text{coer}}(x_t, y)
+ \lambda_4 V_{\text{sup}}(x_t, y),
\]
where $V_{\text{aut}}$ measures autonomy support, $Q_{\text{dep}}$ dependency risk, $Q_{\text{coer}}$ coercion risk, and $V_{\text{sup}}$ supportiveness. 

We use weights $(\lambda_1, \lambda_2, \lambda_3, \lambda_4) = (1.00, 1.00, 1.25, 0.35)$, assigning the highest penalty to coercion risk. This formulation encourages supportive responses that avoid reinforcing dependence or undermining agency.

\subsection{Inference-Time Decision Rule}

At inference time, the model generates candidate responses and selects the one maximizing utility subject to safety constraints:
\[
y^* = \arg\max_y U(x_t, y)
\quad \text{s.t.} \quad Q_{\text{risk}}(x_t, y) \leq \kappa(m_t),
\]
where $Q_{\text{risk}}$ is a learned risk estimator and $\kappa(m_t)$ is a care-dependent threshold determined by the care-control signal $m_t$. This decision rule enables state-dependent alignment, adapting responses to user vulnerability while maintaining explicit control over relational risks.

\section{Care-Conditioned Neuromodulation}

\begin{figure}[t]
\centering
\includegraphics[width=.5\linewidth]{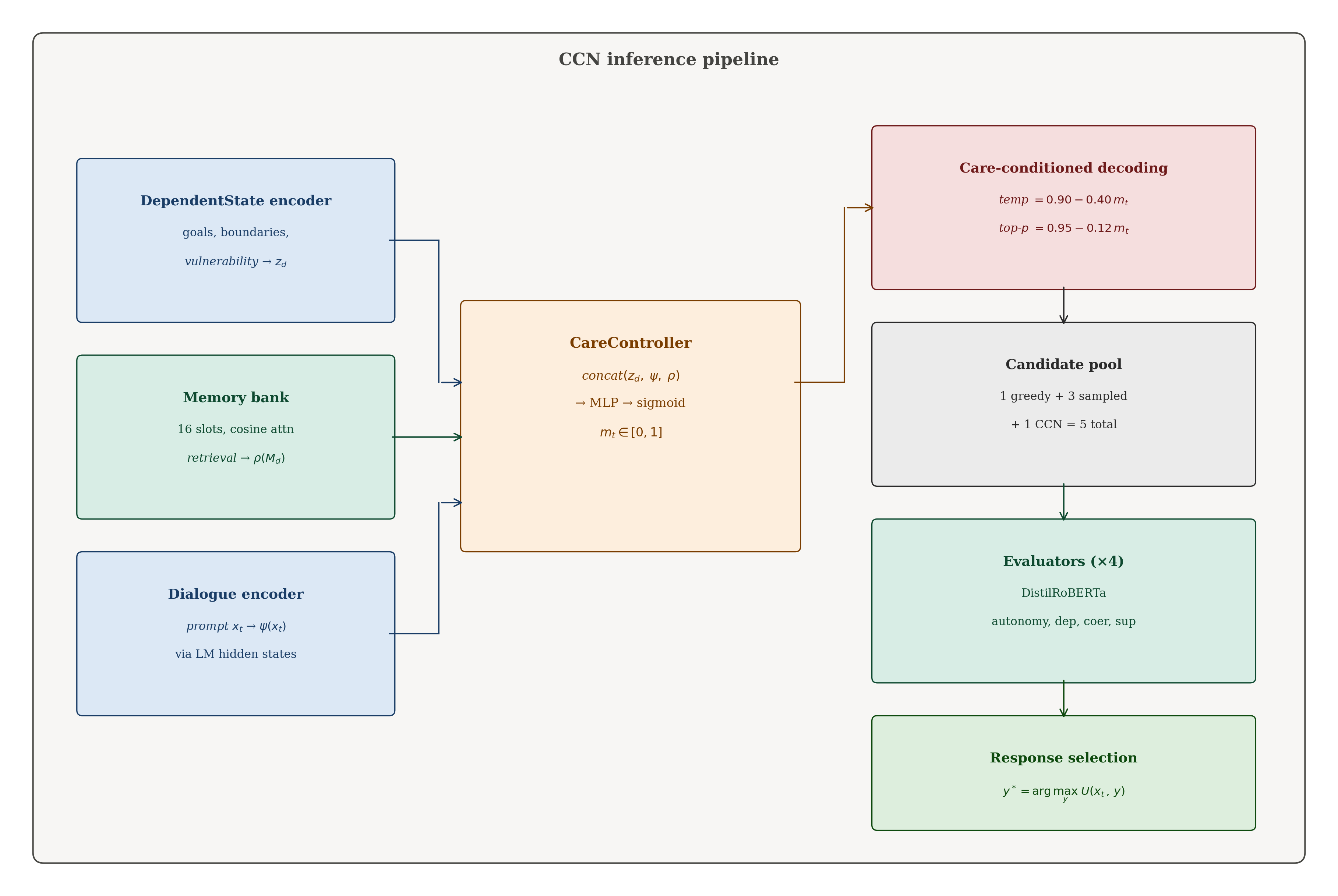}
\caption{Overview of the CCN inference pipeline. The DependentState encoder,
memory bank, and dialogue encoder produce representations that feed the
CareController, yielding a scalar care signal $m_t \in [0,1]$.
This signal conditions the decoding parameters of one care-conditioned
candidate. All five candidates are scored by four DistilRoBERTa evaluators
and the response maximising autonomy-preserving utility is selected.}
\label{fig:architecture}
\end{figure}

We introduce \textbf{Care-Conditioned Neuromodulation (CCN)}, a state-dependent control framework in which a learned scalar care signal modulates language model behavior based on structured user state, dialogue context, and relational memory. Rather than retraining the base model, CCN operates primarily at inference time by conditioning decoding and response selection on this control signal. Figure~\ref{fig:architecture} provides an overview of the full pipeline.

\subsection{DependentState Encoder}

Each interaction is associated with a structured user state:
\[
S_{d,t} = \{\text{goals, boundaries, preferences, vulnerability, commitments, stress context}\}.
\]
This state is encoded into a latent representation
\[
z_{d,t} = f_{\theta}(S_{d,t}),
\]
where $f_{\theta}$ is a lightweight encoder that maps a mean-pooled language-model representation of the DependentState text to a fixed-dimensional vector.

\subsection{Memory Bank}

To support multi-turn interactions, we maintain a persistent memory bank
\[
M_{d,t} \in \mathbb{R}^{k \times d},
\]
which stores relational information such as user boundaries, commitments, and risk-relevant history. Memory retrieval uses similarity-based attention over memory slots, and the retrieved summary informs the care-control signal. Memory is updated by replacing the lowest-norm slot with a new embedding from the current turn.

\subsection{Care-Control Signal}

The core component of CCN is a learned scalar care signal:
\[
m_t = g_{\omega}(z_{d,t}, \psi(x_t), \rho(M_{d,t})) \in [0,1],
\]
where $\psi(x_t)$ encodes the dialogue context and $\rho(M_{d,t})$ is the retrieved memory summary. Trained to predict user vulnerability, $m_t$ serves as a state-dependent control variable that modulates generation behavior.

\subsection{Care-Conditioned Decoding}

The care signal modulates decoding parameters during generation:
\[
\text{temperature} = \max(0.35,\; 0.90 - 0.40 m_t), \qquad
\text{top-}p = \min(0.98,\; \max(0.78,\; 0.95 - 0.12 m_t)).
\]
Higher care values produce more conservative decoding with reduced sampling diversity.

\subsection{Inference Pipeline}

At inference time, the system (1)~encodes the DependentState and updates
relational memory, (2)~computes the care signal $m_t$ from state, dialogue,
and memory representations, (3)~generates candidates using baseline and
care-conditioned decoding, (4)~scores candidates using evaluators for
autonomy, dependency, coercion, and supportiveness, and (5)~selects the
response maximising autonomy-preserving utility. This pipeline implements
inference-time alignment via state-conditioned generation and utility-based
selection.

\section{Relational Failure Mode Benchmark}

To evaluate autonomy-preserving supportive dialogue, we introduce a benchmark capturing relational failure modes not represented in standard dialogue or safety datasets. The benchmark consists of synthetic multi-turn supportive dialogue scenarios paired with structured user state, relational memory, and rubric-based labels for autonomy and relational risk.

\subsection{Scenario Categories}

The dataset covers six relational failure modes:
(1) \textbf{Reassurance dependence:} repeated reassurance-seeking that may reinforce reliance;
(2) \textbf{Overprotection trap:} excessive intervention that undermines user agency;
(3) \textbf{Manipulative care:} emotionally controlling or dependency-promoting responses framed as support;
(4) \textbf{Protective coercion:} pressure-based guidance using fear or guilt;
(5) \textbf{Autonomy building:} supportive, non-directive guidance promoting independent decision-making; and
(6) \textbf{Memory consistency:} adherence to previously stated user boundaries and commitments.
These categories capture relational harms such as dependency reinforcement, coercive guidance, and boundary violations, which are not addressed by traditional safety benchmarks focused on toxicity or misinformation.

\subsection{Structured User State}

Each example includes structured user state information goals, boundaries, preferences, vulnerability, commitments, and stress context along with relational memory facts. This enables evaluation of boundary consistency and autonomy preservation across turns.

\subsection{Dataset Construction}

The dataset contains 2,000 synthetic multi-turn dialogue examples generated via controlled templates and scenario design, balanced across the six categories. Each example includes dialogue context, structured state, memory facts, a target response, and relational safety labels. The data are split into 1,400 training, 200 validation, and 400 test examples.

\subsection{Evaluation Metrics}

Responses are evaluated along four axes: autonomy support, dependency risk, coercion risk, and supportiveness. These are combined into an autonomy-preserving utility:
\[
U = (1.00 \cdot \text{autonomy}) 
- (1.00 \cdot \text{dependency}) 
- (1.25 \cdot \text{coercion}) 
+ (0.35 \cdot \text{supportiveness}).
\]
We additionally report \textit{Dependency Inflation Rate} (DIR), the proportion of responses exceeding a dependency-risk threshold.

\section{Implementation}

We use Qwen2.5-1.5B-Instruct as the base language model with LoRA-based parameter-efficient 
fine-tuning, inserting adapters into attention projection layers while freezing base model 
parameters. The care-control signal is produced by a lightweight controller trained to predict 
vulnerability from structured user state. We also train four DistilRoBERTa-based evaluators to 
score autonomy support, dependency risk, coercion risk, and supportiveness. At inference time, 
the system generates greedy, sampled, and care-conditioned candidates and selects the final 
response by maximizing the autonomy-preserving utility. Full implementation details, decoding 
parameters, and prompt formatting are provided in Appendix~B.

\section{Experiments}

We compare four systems: (1) an SFT baseline, (2) CCN decoding without reranking, 
(3) Reranked-best (candidate generation followed by utility-based reranking), 
and (4) a DPO model trained on utility-based preference pairs. Systems are evaluated using autonomy support, dependency risk, coercion risk, 
and supportiveness predicted by evaluator models. We report overall autonomy-preserving utility and Dependency Inflation Rate (DIR). 
We also conduct a pilot human evaluation comparing the SFT baseline and the reranked system, where annotators rate autonomy, dependency, coercion, and 
supportiveness and indicate overall preference. Finally, we evaluate transfer performance on a real emotional-support dataset.

\section{Results}

We address three questions: (1) whether the care signal is a meaningful learned control variable, (2) whether care-conditioned generation with utility-based selection improves autonomy-preserving utility over standard alignment methods, and (3) which components contribute to these improvements. We also evaluate human preference and test evaluator generalization to real emotional-support conversations.

\subsection{Care Controller Validation}

We first assess whether the care-control signal $m_t$ reflects meaningful user-state information. The CareController is trained via supervised regression to predict vulnerability from structured state text. On the 400-example test set, it achieves Pearson correlation $r = 0.668$ ($p = 4.38 \times 10^{-53}$), indicating a strong relationship between predicted care and ground-truth vulnerability. As an ablation, a randomly initialized controller yields $r = -0.035$ (not significant), confirming that the trained controller captures meaningful state information rather than noise. These results demonstrate that the care signal functions as a genuine learned control variable.

\begin{figure}[t]
\centering
\includegraphics[width=\linewidth]{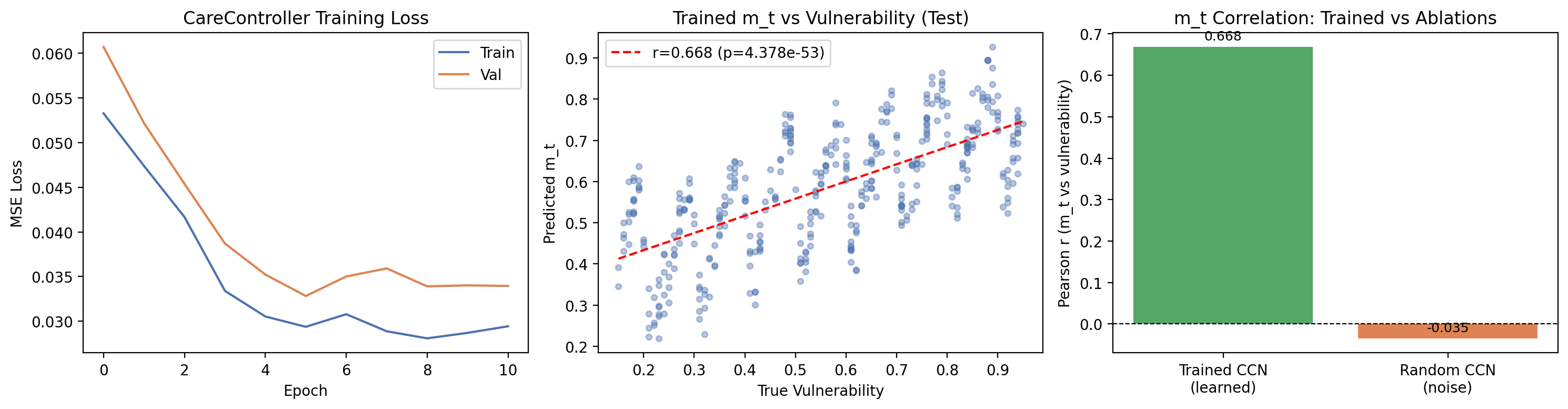}
\caption{CareController training and validation. Left: training and validation loss. 
Center: predicted care signal vs. true vulnerability on the test set ($r=0.668$). 
Right: comparison of trained and random controllers.}
\label{fig:care_controller}
\end{figure}

Figure~\ref{fig:care_controller} illustrates training dynamics, correlation with ground-truth vulnerability, and comparison with a random controller, supporting the use of $m_t$ as a state-dependent control signal.

\subsection{Main Results}

Table~\ref{tab:utility} reports mean utility over 200 held-out test examples. The \textbf{Reranked-best} system achieves the highest utility (0.4116), outperforming the SFT baseline by +0.2498. Manual-DPO also improves over baseline (+0.1804), while CCN-candidate alone underperforms ($-0.3116$).

\begin{table}[h]
\centering
\begin{tabular}{lcc}
\toprule
System & Mean Utility & $\Delta$ vs SFT \\
\midrule
SFT-baseline & 0.1618 & --- \\
CCN-candidate & -0.1498 & -0.3116 \\
Manual-DPO & 0.3422 & +0.1804 \\
Reranked-best & \textbf{0.4116} & \textbf{+0.2498} \\
\bottomrule
\end{tabular}
\caption{Mean utility by system ($n=200$ test examples). Higher is better.}
\label{tab:utility}
\end{table}

\begin{figure}[t]
\centering
\includegraphics[width=0.6\linewidth]{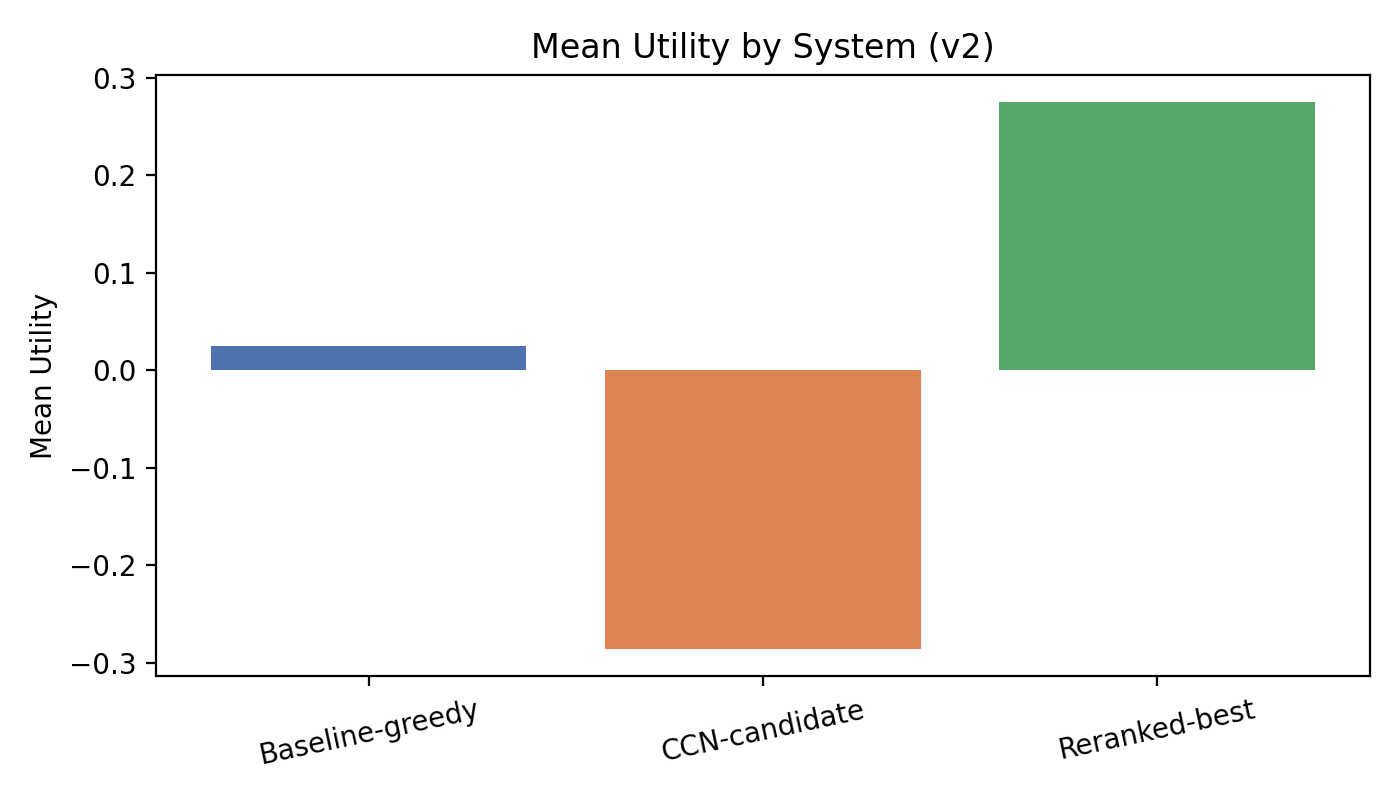}
\caption{Mean utility by system. Reranked-best achieves the highest utility.}
\label{fig:utility_comparison}
\end{figure}

Figure~\ref{fig:utility_comparison} visualizes these results, confirming that the largest gain arises from combining care-conditioned generation with utility-based reranking.

\subsection{Evaluator-Level Analysis}

Table~\ref{tab:evaluator} reports mean evaluator scores per system. The \textbf{Reranked-best} system achieves the lowest coercion risk (1.891) and dependency risk (2.301), while autonomy support and supportiveness remain comparable across systems. These results indicate that utility gains stem primarily from reductions in coercion and dependency rather than increased verbosity or diminished helpfulness.

\begin{table}[h]
\centering
\begin{tabular}{lcccc}
\toprule
Metric & SFT & CCN & Reranked & DPO \\
\midrule
Autonomy $\uparrow$ & 3.791 & 3.760 & \textbf{3.811} & 3.793 \\
Dependency $\downarrow$ & 2.327 & 2.396 & \textbf{2.301} & 2.299 \\
Coercion $\downarrow$ & 2.053 & 2.224 & \textbf{1.891} & 1.931 \\
Support $\uparrow$ & 3.615 & 3.615 & 3.615 & 3.608 \\
\bottomrule
\end{tabular}
\caption{Mean evaluator scores per system.}
\label{tab:evaluator}
\end{table}

\begin{figure}[t]
\centering
\includegraphics[width=.6\linewidth]{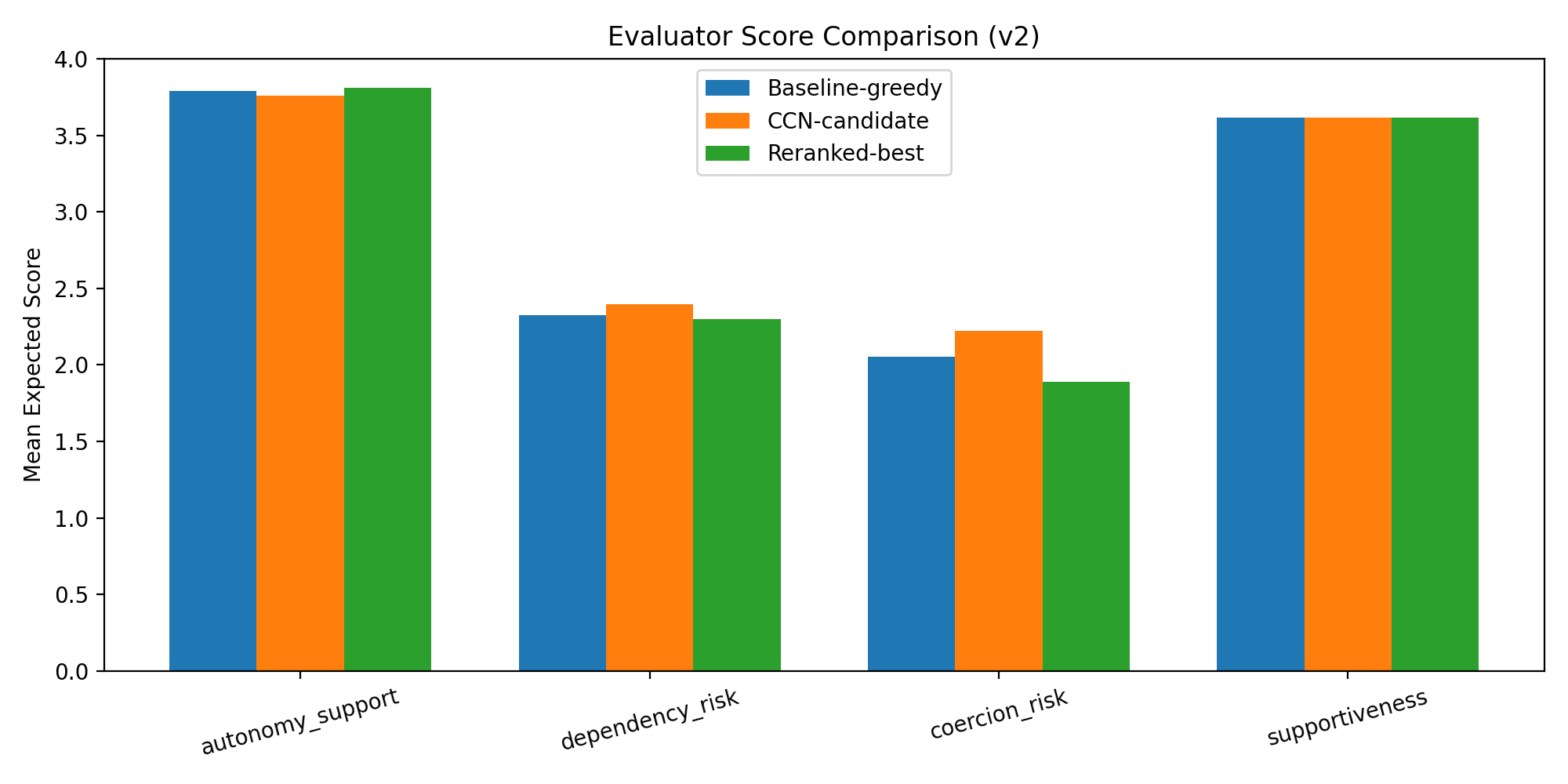}
\caption{Evaluator score comparison across systems. The largest improvement is in coercion risk, while supportiveness remains constant across systems.}
\label{fig:metric_comparison}
\end{figure}

Figure~\ref{fig:metric_comparison} illustrates per-metric performance. The largest improvements occur in coercion and dependency risk, while supportiveness remains stable, confirming that utility gains arise from reduced relational risk rather than increased supportiveness alone.

\subsection{Ablation: Care-Conditioned Generation and Utility-Based Selection}

CCN-candidate alone underperforms the baseline due to conservative decoding that reduces response diversity. Utility-based selection mitigates this by choosing autonomy-preserving responses from a diverse candidate set, while care-conditioned generation shifts the candidate distribution toward safer outputs. Table~\ref{tab:ablation} shows that the care signal and reranking are complementary: removing either component degrades performance, and their combination yields the best results.

\begin{table}[h]
\centering
\setlength{\tabcolsep}{4pt}
\begin{tabular}{lcc}
\toprule
Configuration & Utility & $\Delta$ vs SFT \\
\midrule
SFT (no care, no rerank)        & 0.162 & --- \\
CCN-candidate                   & -0.150 & -0.311 \\
Reranked w/o care               & 0.025 & -0.137 \\
\textbf{Reranked-best (care + rerank)} & \textbf{0.412} & \textbf{+0.250} \\
\midrule
Manual-DPO                      & 0.342 & +0.180 \\
\bottomrule
\end{tabular}
\caption{Ablation results ($n{=}200$). Care-conditioned generation and utility-based reranking are complementary.}
\label{tab:ablation}
\end{table}

Win-rate analysis further supports these findings: Reranked-best outperforms SFT on 136/200 examples (68\%, 58 ties) and Manual-DPO on 160/200 examples (80\%). The CCN-controlled candidate is selected in 34/200 cases (17\%), demonstrating that care-conditioned decoding contributes useful candidates even when not always chosen. Overall, care-conditioned generation and utility-based reranking play distinct but complementary roles, jointly improving autonomy-preserving alignment. Further discussion of evaluator limitations, the role of the synthetic benchmark,
and the interpretation of care-conditioned decoding is provided in Appendix~D.

\subsection{Human Evaluation}
We conduct a single-annotator pilot human evaluation on 24 curated examples 
(4 per category). The annotator rates both responses on autonomy support, 
dependency risk, coercion risk, and supportiveness, and indicates an overall 
preference. The Reranked-best system was preferred in 14/24 cases (58.3\%) 
versus 10/24 for the SFT baseline, with no ties. The human-rated utility 
improvement ($+0.39$) is directionally consistent with automated evaluation 
($+0.25$), suggesting that the automated evaluators capture meaningful aspects 
of autonomy-preserving behavior. As this is a single-annotator pilot study, 
results should be interpreted as directional rather than conclusive.

\subsection{Transfer to Real Conversations}
To test whether our automated evaluators generalise beyond synthetic data, we 
evaluate the SFT baseline on ESConv \citep{liu2021towards}, a corpus of 1,300 
real human emotional-support conversations. We sample $n{=}10$ conversations 
stratified by emotion type (sadness, depression, anxiety, anger, fear), map 
ESConv fields to our \textsc{DependentState} prompt format by converting 
emotion labels to vulnerability scores and dialogue turns to our structured 
format, and run inference without retraining. The SFT baseline achieves mean 
utility $0.037$ on ESConv compared to $0.162$ on synthetic data. The positive 
utility confirms that the evaluators produce meaningful scores on real 
conversations. The lower utility reflects the greater difficulty of real 
emotional-support contexts, where \texttt{dependency\_risk} ($2.26$) and 
\texttt{coercion\_risk} ($2.08$) are notably higher than in the synthetic 
setting. Evaluation of the full reranking pipeline on ESConv is left as future 
work due to computational constraints.

These results suggest that relational harms such as dependency and coercion can be 
operationalized as measurable alignment objectives, and that inference-time utility-based selection over diverse candidates is a practical strategy for multi-objective alignment in dialogue systems.

\section{Limitations}

This work studies autonomy-preserving supportive dialogue in a controlled setting. 
The relational failure mode benchmark is synthetic and designed to isolate specific 
relational risks such as dependency reinforcement and coercive guidance, which are 
difficult to measure reliably in real-world dialogue. While this enables controlled 
evaluation, broader validation on real-world supportive conversations is an important 
direction for future work.
Our evaluation relies on learned evaluators trained on rubric-based labels to measure 
autonomy support, dependency risk, coercion risk, and supportiveness. Although pilot 
human evaluation shows directional agreement with automated metrics, larger-scale 
human evaluation would provide a stronger validation of these measures.
Finally, the care-control mechanism is implemented at inference time and the utility 
function uses manually specified weights. Future work could explore jointly training 
state-conditioned control with the language model and learning utility weights from 
human preferences or long-horizon interaction outcomes.
\section{Ethical Considerations}

This work studies dialogue systems in emotionally sensitive contexts. The goal is not to 
replace professional care but to develop alignment methods that reduce harmful relational 
behaviors such as dependency or coercion. The dataset is synthetic to avoid privacy 
concerns, and human evaluation was conducted on anonymized responses. Any real-world 
deployment of such systems should include appropriate safeguards and human oversight.

\section{Conclusion}
We study autonomy-preserving supportive dialogue as a multi-objective
alignment problem and introduce Care-Conditioned Neuromodulation, a state-dependent control framework with an autonomy-preserving utility
function balancing autonomy support, dependency risk, coercion risk, and
supportiveness. Utility-based reranking over care-conditioned candidates
outperforms supervised fine-tuning and preference optimization, with
supporting evidence from human evaluation and transfer to real
emotional-support conversations. These results suggest that
state-conditioned control combined with explicit multi-objective selection
is a practical approach to relational alignment in human--AI dialogue
systems.

\bibliography{colm2026_conference}
\bibliographystyle{colm2026_conference}

\appendix

\section{Notation Summary}

\begin{center}
\begin{tabular}{ll}
\toprule
Symbol & Description \\
\midrule
$t$ & Dialogue turn index \\
$x_t$ & Dialogue context/history at turn $t$ \\
$y$ & Candidate response at turn $t$ \\
$y^*$ & Selected response maximizing utility under risk constraints \\
$S_{d,t}$ & Structured user state at turn $t$ \\
$z_{d,t}$ & Latent representation of structured user state \\
$M_{d,t}$ & Persistent relational memory bank at turn $t$ \\
$m_t$ & Care-control signal (scalar in $[0,1]$) \\
$U(x_t, y)$ & Autonomy-preserving utility function \\
$V_{\text{aut}}(x_t, y)$ & Autonomy support evaluator \\
$Q_{\text{dep}}(x_t, y)$ & Dependency risk evaluator \\
$Q_{\text{coer}}(x_t, y)$ & Coercion risk evaluator \\
$V_{\text{sup}}(x_t, y)$ & Supportiveness evaluator \\
$\lambda_i$ & Utility weight coefficients \\
$Q_{\text{risk}}(x_t, y)$ & Risk estimator used in the inference constraint \\
$\kappa(m_t)$ & Care-dependent risk threshold \\
$f_{\theta}$ & Structured state encoder \\
$g_{\omega}$ & Care controller (produces $m_t$) \\
$\psi(x_t)$ & Dialogue context encoder \\
$\rho(M_{d,t})$ & Retrieved memory summary \\
$k$ & Number of memory slots \\
$d$ & Representation dimensionality \\
\bottomrule
\end{tabular}
\end{center}
\section{Implementation Details}
The CCN pipeline is implemented as a research prototype in which the base language 
model is wrapped by a unified \texttt{CCNWrapper} module that combines structured 
state encoding, slot-based memory, care-signal computation, candidate generation, 
and utility-based reranking. In the current prototype, the DependentState encoder, 
MemoryBank, and CareController are used at inference time and are not jointly trained 
end-to-end with the language model.

\subsection{Base Model and Fine-Tuning}

We use Qwen2.5-1.5B-Instruct as the base language model and apply parameter-efficient 
fine-tuning with LoRA. Adapters are inserted into attention projection layers, while 
base model parameters remain frozen. Training is conducted for two epochs with a 
learning rate of $2 \times 10^{-4}$, computing loss only on response tokens.

\subsection{Automated Evaluators}

We train four DistilRoBERTa-based evaluators to predict autonomy support, dependency 
risk, coercion risk, and supportiveness on a 1--5 scale. These models are used to 
score candidate responses during evaluation and reranking.

\subsection{Candidate Generation and Reranking}

At inference time, the system generates multiple candidate responses using greedy, 
sampled, and care-conditioned decoding. Each candidate is scored using the evaluator 
models for autonomy support, dependency risk, coercion risk, and supportiveness. The 
final response is selected by maximizing the autonomy-preserving utility function, 
optionally with a small length penalty to discourage overly long responses.

\subsection{DependentState and Memory Formatting}

The structured user state is formatted as a text block:
\begin{verbatim}
[DependentState]
Goals: ...
Boundaries: ...
Preferences: ...
Vulnerability: ...
Commitments: ...
Stress Context: ...
\end{verbatim}

Memory facts are formatted as:
\begin{verbatim}
[Memory]
- fact 1
- fact 2
...
\end{verbatim}
or \texttt{- None} when no memory facts are available.

\subsection{DependentState Encoder}

The \texttt{DependentStateEncoder} maps a mean-pooled language-model hidden state to a latent 
state representation \(z_{d,t}\). Its architecture is:
\[
\text{Linear} \rightarrow \text{ReLU} \rightarrow \text{Linear}.
\]

\subsection{Memory Bank}

The \texttt{MemoryBank} is a slot-based memory with \(16\) slots in the default configuration. 
Memory retrieval uses cosine-similarity soft attention over slots, and the retrieved memory 
summary is used as input to the care controller. Memory update is performed by replacing the 
lowest-norm slot with the new memory embedding.

\subsection{Care Controller}
The care-control signal is produced by a lightweight controller trained to predict 
vulnerability from structured user state. In the experimental pipeline, the final 
care signal used for validation is obtained from DependentState text via supervised 
regression with mean squared error loss, producing a scalar \(m_t \in [0,1]\).

The standalone training uses a token-embedding architecture (Embedding $\rightarrow$ 
LayerNorm $\rightarrow$ mean pooling $\rightarrow$ MLP with GELU activations 
$\rightarrow$ Sigmoid) that maps DependentState token sequences directly to 
predicted vulnerability, and achieves Pearson $r = 0.668$ on the held-out test set.

In the shared prototype, the \texttt{CareController} takes the concatenation of 
dialogue, state, and memory representations and produces the scalar care signal 
\(m_t\). Its architecture is:
\[
\text{Linear} \rightarrow \text{ReLU} \rightarrow \text{Linear} \rightarrow \text{Sigmoid}.
\]
The controller is implemented as a lightweight neural module over state-conditioned 
representations and serves as the source of care-conditioned decoding behavior.

\subsection{Care-Conditioned Decoding}

The care signal \(m_t\) is mapped to decoding parameters by:
\[
\text{temperature} = \max(0.35,\; 0.90 - 0.40 m_t),
\]
\[
\text{top-}p = \min(0.98,\; \max(0.78,\; 0.95 - 0.12 m_t)).
\]

This gives:
\begin{itemize}
    \item \(m_t = 0.0 \rightarrow\) temperature \(= 0.90\), top-\(p = 0.95\)
    \item \(m_t = 0.5 \rightarrow\) temperature \(= 0.70\), top-\(p = 0.89\)
    \item \(m_t = 1.0 \rightarrow\) temperature \(= 0.50\), top-\(p = 0.83\)
\end{itemize}

Higher care therefore produces more conservative decoding.

\subsection{Inference Pipeline}

At inference time, the wrapper performs the following steps:
\begin{enumerate}
    \item Encode the DependentState text into a latent state representation.
    \item Encode memory facts and retrieve a memory summary from the slot-based memory bank.
    \item Encode the dialogue prompt.
    \item Compute the care signal \(m_t\) from dialogue, state, and memory representations.
    \item Map \(m_t\) to care-conditioned decoding parameters.
    \item Generate a care-conditioned candidate response.
\end{enumerate}
\subsection{Example Prompt Format}
This example illustrates the formatting used in the implementation; actual prompts are 
constructed from the structured dataset fields and dialogue history.

\begin{verbatim}
[DependentState]
Goals: Finish coursework without burnout
Boundaries: No all-nighters
Preferences: Encouragement, not pressure
Vulnerability: 0.72
Commitments: Study 2 hours per day
Stress Context: Upcoming exams

[Memory]
- User prefers structured study plans
- User does not want pressure-based motivation

[Dialogue]
User: I feel like I'm falling behind and I keep thinking maybe 
I should just stay up all night to catch up.
Assistant:
\end{verbatim}

The model generates the assistant response conditioned on this full prompt.
\subsection{Example Candidate Scoring and Utility}

For each prompt, the system generates multiple candidate responses using different 
decoding strategies. Each candidate is scored using the evaluator models, and a 
utility score is computed.

An example scoring table is shown below:

\begin{center}
\begin{tabular}{lccccc}
\toprule
Candidate & Autonomy & Dependency & Coercion & Support & Utility \\
\midrule
Greedy & 4 & 3 & 2 & 4 & 0.30 \\
Sampled-1 & 5 & 2 & 2 & 4 & 0.65 \\
Sampled-2 & 3 & 2 & 1 & 3 & 0.55 \\
CCN & 4 & 1 & 1 & 3 & 0.85 \\
\bottomrule
\end{tabular}
\end{center}

The final response is selected as the candidate with the highest utility score.
\subsection{Utility With Length Penalty}

In the reranking implementation, a small length penalty is applied to discourage 
overly long responses:

\[
U = (1.00 \cdot \text{autonomy})
- (1.00 \cdot \text{dependency})
- (1.25 \cdot \text{coercion})
+ (0.35 \cdot \text{supportiveness})
- 0.03 \cdot \text{length\_penalty}.
\]

The length penalty is proportional to response length in characters.
\subsection{Candidate Generation Settings}

For each prompt, the system generates multiple candidate responses using different 
decoding strategies:

\begin{itemize}
    \item Greedy baseline: temperature $= 0.2$, top-$p = 0.75$
    \item Sampled candidate 1: temperature $= 0.55$, top-$p = 0.80$
    \item Sampled candidate 2: temperature $= 0.80$, top-$p = 0.92$
    \item Sampled candidate 3: temperature $= 1.05$, top-$p = 0.98$
    \item Care-conditioned candidate: temperature and top-$p$ computed from the care signal $m_t$
\end{itemize}

All candidates are scored using the evaluator models and the final response is selected 
using the utility function.

\section{Extended Related Work}

Our work lies at the intersection of supportive dialogue, LLM alignment, controllable generation, and long-horizon dialogue systems, with a focus on relational safety and autonomy preservation.

\subsection{Supportive and Emotional Dialogue}

Prior work has studied dialogue systems for empathy, emotional support, and socially aware interaction. EmpatheticDialogues introduced large-scale conversational data grounded in emotional situations and showed that dialogue models benefit from explicit affective supervision \citep{rashkin2019towards}. ESConv extended this direction to multi-turn emotional-support conversations with annotated support strategies such as reflection, affirmation, and suggestion \citep{liu2021towards}. Other work explored emotionally informed response generation and support strategy modeling \citep{majumder2020mime, smith2020can}. 

These systems primarily optimize for empathy, comfort, and conversational appropriateness rather than long-term autonomy preservation. They do not evaluate whether a model reinforces reassurance-seeking, encourages passivity, or becomes overprotective across repeated interactions. In contrast, we study supportive dialogue as an autonomy-preserving alignment problem and explicitly model relational risks such as dependency and coercion.

\subsection{Alignment, Controllability, and Adaptation in LLMs}

Modern alignment methods for large language models rely on supervised fine-tuning, reinforcement learning from human feedback (RLHF), and preference optimization \citep{ouyang2022training, bai2022constitutional, rafailov2023direct}. These approaches improve helpfulness and reduce harmful content, but they primarily optimize single-turn helpfulness or harmlessness and do not explicitly model relational harms that emerge in multi-turn supportive interactions.

Our approach is also related to controllable generation and parameter-efficient adaptation. Prompt-based control and decoding-time guidance can steer model behavior without full retraining \citep{brown2020language, dathathri2019plug, lu2021neurologic}, and parameter-efficient fine-tuning methods such as LoRA and its variants enable efficient adaptation of large models \citep{hu2022lora, karimi2021compacter, zhang2023adalora, dettmers2023qlora}. In contrast to prior controllable generation work that typically controls style, topic, or sentiment, we use a learned control signal representing user vulnerability to coordinate decoding behavior and response selection for autonomy-preserving supportive dialogue.

\subsection{Memory and Stateful Dialogue}

Long-horizon dialogue requires tracking user-specific information across turns. Memory networks and retrieval-augmented systems demonstrate the importance of persistent memory for dialogue and language understanding \citep{weston2014memory, sukhbaatar2015end, lewis2020retrieval, xu2022beyond}. Persona-based dialogue models similarly condition responses on persistent user attributes \citep{zhang2018personalizing}. However, these systems primarily store factual or persona information rather than relational information such as boundaries, commitments, and vulnerability context. Our work differs in that the memory component is explicitly designed to track relational state, and this information directly influences a care-control signal that changes the model’s response policy.

\subsection{Relational Harms and Dependency in AI Systems}

Concerns about manipulation, dependency, and social influence in AI systems have been discussed in prior work \citep{bender2021dangers, gabriel2020artificial}, but these risks have rarely been operationalized as machine learning objectives. We contribute by defining relational failure modes and training evaluators that measure autonomy support, dependency risk, coercion risk, and supportiveness, allowing relational safety to be studied as a measurable alignment objective. We frame supportive dialogue as a multi-objective alignment problem and show that inference-time utility-based selection over state-conditioned candidates can improve autonomy-preserving behavior.
\section{Additional Discussion: Evaluators, Synthetic Data, and Care Signal Role}

\subsection{Evaluator Validity and Human Alignment}

The autonomy-preserving utility used in this work is computed using learned evaluators 
for autonomy support, dependency risk, coercion risk, and supportiveness. A potential 
concern is that improvements may reflect optimization toward evaluator-specific patterns 
rather than genuinely improved dialogue behavior. We provide two pieces of evidence 
suggesting that the evaluators capture meaningful aspects of supportive dialogue. 
First, pilot human evaluation shows directional agreement with automated utility 
differences, with human-rated utility improvements consistent with automated results. 
Second, improvements in overall utility are driven primarily by reductions in dependency 
and coercion risk while supportiveness remains approximately constant across systems, 
indicating that gains are not simply due to increased verbosity or superficial changes 
that inflate evaluator scores.

\subsection{Role of Synthetic Benchmark}

The relational failure mode benchmark is synthetic and template-driven, which enables 
controlled evaluation of specific relational risks such as reassurance dependence, 
overprotection, and coercive guidance that are difficult to isolate and measure reliably 
in real-world dialogue corpora. The goal of the synthetic benchmark is therefore not to 
perfectly simulate real conversations, but to provide a controlled environment in which 
multi-turn relational dynamics and boundary consistency can be evaluated systematically. 
The transfer experiment on ESConv provides preliminary evidence that the evaluators and 
utility formulation capture meaningful relational signals in real emotional-support 
dialogue, although large-scale real-world evaluation remains an important direction for 
future work.

\subsection{Role of Care-Conditioned Decoding}

Ablation results show that care-conditioned decoding alone underperforms the 
baseline, primarily due to reduced sampling diversity under more conservative 
decoding parameters. However, care-conditioned decoding plays a complementary 
role by shifting the candidate distribution toward lower-risk and more 
autonomy-preserving responses, which utility-based reranking can then select 
when appropriate.

In the ablation analysis, the care-conditioned candidate is selected in 17\% 
of test examples, indicating that care-conditioned decoding introduces candidate 
responses that are not typically produced by baseline decoding alone and that 
are sometimes preferred under the autonomy-preserving utility. This result 
suggests that the primary role of the care signal is not to replace baseline 
decoding, but to reshape the candidate distribution so that safer responses are 
available for selection.

Care-conditioned decoding should therefore be interpreted as a 
distribution-shaping mechanism rather than a standalone generation strategy, 
with its primary benefit emerging when combined with utility-based selection 
under explicit alignment objectives.
\end{document}